%% file: main.tex
\documentclass[10pt,twocolumn,letterpaper]{article}

\usepackage{iccv}
\usepackage{times}
\usepackage{epsfig}
\usepackage{graphicx}
\usepackage{amsmath}
\usepackage{xcolor}
\usepackage{amssymb}
\usepackage{textcomp}
\usepackage{siunitx}
\usepackage[font=small,skip=-1pt]{caption}
\usepackage[pagebackref=true,breaklinks=true,letterpaper=true,colorlinks,bookmarks=false]{hyperref}

\iccvfinalcopy 

\ificcvfinal\fi

\begin{document}


\title{Cross-Domain Image Classification through Neural-Style \\Transfer Data Augmentation}
\author{Yijie Xu\\
University College London\\
{\tt\small yijie.xu.10@ucl.ac.uk}
\and
Arushi Goel\\
University of Edinburgh\\
{\tt\small goel.arushi@gmail.com}
}

\maketitle
\ificcvfinal\thispagestyle{empty}\fi

\begin{abstract}
In particular, the lack of sufficient amounts of domain-specific data can reduce the accuracy of a classifier. In this paper, we explore the effects of style transfer-based data transformation on the accuracy of a convolutional neural network classifiers in the context of automobile detection under adverse winter weather conditions. The detection of automobiles under highly adverse weather conditions is a difficult task as such conditions present large amounts of noise in each image. The InceptionV2 architecture is trained on a composite dataset, consisting of either normal car image dataset , a mixture of normal and style transferred  car images, or a mixture of normal car images and those taken at blizzard conditions, at a ratio of 80:20. All three classifiers are then tested on a dataset of  car images taken at blizzard conditions and on vehicle-free snow landscape images. We evaluate and contrast the effectiveness of each classifier upon each dataset, and discuss the strengths and weaknesses of style-transfer based approaches to data augmentation.

\end{abstract}
\section{Introduction}
\label{sec.intro}
\input{intro}

\section{Related Work}
\label{sec.rel}
\input{rel}

\section{Proposed Method}
In this section, we describe our method to achieve data augmentation using neural-style transfer.
\label{sec.method}
\input{method}

\section{Results and Experiments}
\label{sec.exp}
\input{exp}

\section{Conclusion}
\label{sec.conclusion}
\input{conclusion}

{\small
\bibliographystyle{ieee}
\bibliography{short}
}

\end{document}

%% file: intro.tex
It is commonly accepted that large amounts of data is needed to train deep neural networks which yields high performance. When the number of training examples is reduced in size, such networks become more prone to overfitting which reduces their generalizable capability \cite{liu2015very}. To improve generalization and performance, several data augmentation and regularization techniques have been explored in the research community \cite{krizhevsky2012imagenet, perez2017effectiveness}. In this paper, we present a data augmentation technique for improving the performance of neural networks. The availability of data is known to improve performance as more  discriminative features  are extracted by the model . However, the resource intensive nature of data acquisition can render it unfeasible, necessitating a need to generate synthetic data.

In this paper, we introduce a novel neural style transfer-based data transformation and augmentation method for cross-domain classification applications, where an existing data domain (cars) is shifted to target a separate cross-domain application (cars obscured with snow).  Due to lack of suitable conditions required to collect data specific to a domain, style-transfer based data augmentation may become a necessary aspect of data collection. In other words, while many applications have standardized  data generation procedures, others are subject to  time, temperature, or cost-sensitive conditions, greatly restricting the data collection time.

One practical example to consider is that of car detection under adverse winter conditions. The acquisition of such data may only be done in geographically suitable areas under suitable weather conditions, which is subject to change without notice. With increasing global temperatures due to climate change, tracking suitable time for data collection is becoming increasingly difficult. 

Our style-transfer based data transformation tackles this problem  by  applying appropriate low-level transformations to existing vanilla car image datasets \cite{gatys2015neural}. As the degree and type of style transfer is controlled by user-defined weights and reference images respectively, the various kinds of synthetic composite outputs could be generated to match a particular feature domain of interest. To the best of our knowledge, we are the first to use neural style transfer for data augmentation to improve cross-domain performance using domain-specific noise.

The rest of the paper discusses related work, our proposed method,  experimental results and conclusion.



%% file: rel.tex
\subsection{Data Augmentation}

Data augmentation improves model robustness and sample size \cite{perez2017effectiveness, li2017demystifying}. Traditional augmentation approaches for image classification applications include rotations, translations, zooms, Gaussian noise addition, and mirror flips. 
More advanced methods to data augmentation have been investigated by Wu \etal \cite{wu2015deep}, who argues that data augmentation itself is invaluable in preventing the learning of image-specific artifacts in training datasets and to improve learning performance features that are invariant for specific classes. Besides traditional augmentation methods, they incorporated techniques such as adaptive resolution scaling, color changes, lens distortion effects, and global or localized brightness changes to augment their data, and achieve a performance surpassing the top-scoring models from the ILSVRC classification tasks of the preceding three years \cite{wu2015deep}.

The generation of completely synthetic data as a data augmentation approach has also been investigated for image classification applications, with varying success.  A study by Alhajia \etal \cite{alhaija2017augmented} on automobile classification in traffic situations concluded that a dataset consisting of real images augmented with 3D-modelled vehicles outperformed both a purely 3D-modelled dataset as well as the original dataset \cite{alhaija2017augmented}. Similarly, augmenting clear facial data with synthetically modelled accessories has also been shown to improve recognition performance in surveillance industry \cite{wang2019survey}. More recently, adversarial architectures such as GANs have shown promise in being able to generate new synthetic data for applications such as in medical imaging and facial recognition by combining feature elements from existing data. However, these often suffer from restrictions in feature domains, being subject to domain bias \cite{bowles2018gan, tanaka2019data}. In our method, we target the problem of domain bias by augmenting the training dataset using neural style transfer, which has not yet been explored in existing works.

\subsection{Domain Bias}

The subject of cross-domain classification, or the use of a classifier trained on a particular distribution of data to generalize on other previously unseen datasets, has been investigated in literature. Previous approaches in literature suffered from poor cross-domain discriminative performance, with the phenomenon termed as domain bias, where a model will not generalize well to data not observed during training \cite{uhrig2017sparsity, csurka2017domain}. The differences between such datasets may be ascribed to factors such as variations in camera pose, illumination, lens properties, noise, background, and the presence of foreign artifacts \cite{jackson2018style}.

Transfer learning addresses domain bias by fine-tuning an existing network trained for one domain to classify samples from the other domain \cite{pan2009survey, shao2014transfer, zhang2016robust}. The approach relies on the understanding that convolutional features generated from larger datasets are shared across multiple domains, allowing for a faster and resource-efficient training process to be pursued \cite{saenko2010adapting}. This has been extensively demonstrated for medical applications, where access to large amounts of domain-specific data is limited, the use of network pre-trained on the ImageNet dataset has shown to improve classification accuracy for the detection of mammographic tumours, lung disease, and abnormal lymph nodes in test images \cite{shin2016deep, huynh2016digital}. However, the lack of specifity of such approaches to the target domain has made the acquisition of domain-specific data a preferable course of action.
In contrast, our approach utilizes transfer learning for cross-domain classification using a composite dataset, where the approach consisted of a complete retraining  of the network using  combination of examples from both the original and target domains, wherein the data for the target domain was syntheticially derived from the original dataset using neural style transfer. As such, we aimed for the generation of synthetic data representative of the target domain to improve classifier performance.

\begin{figure*}[t]
\begin{center}
\includegraphics[width=\linewidth]{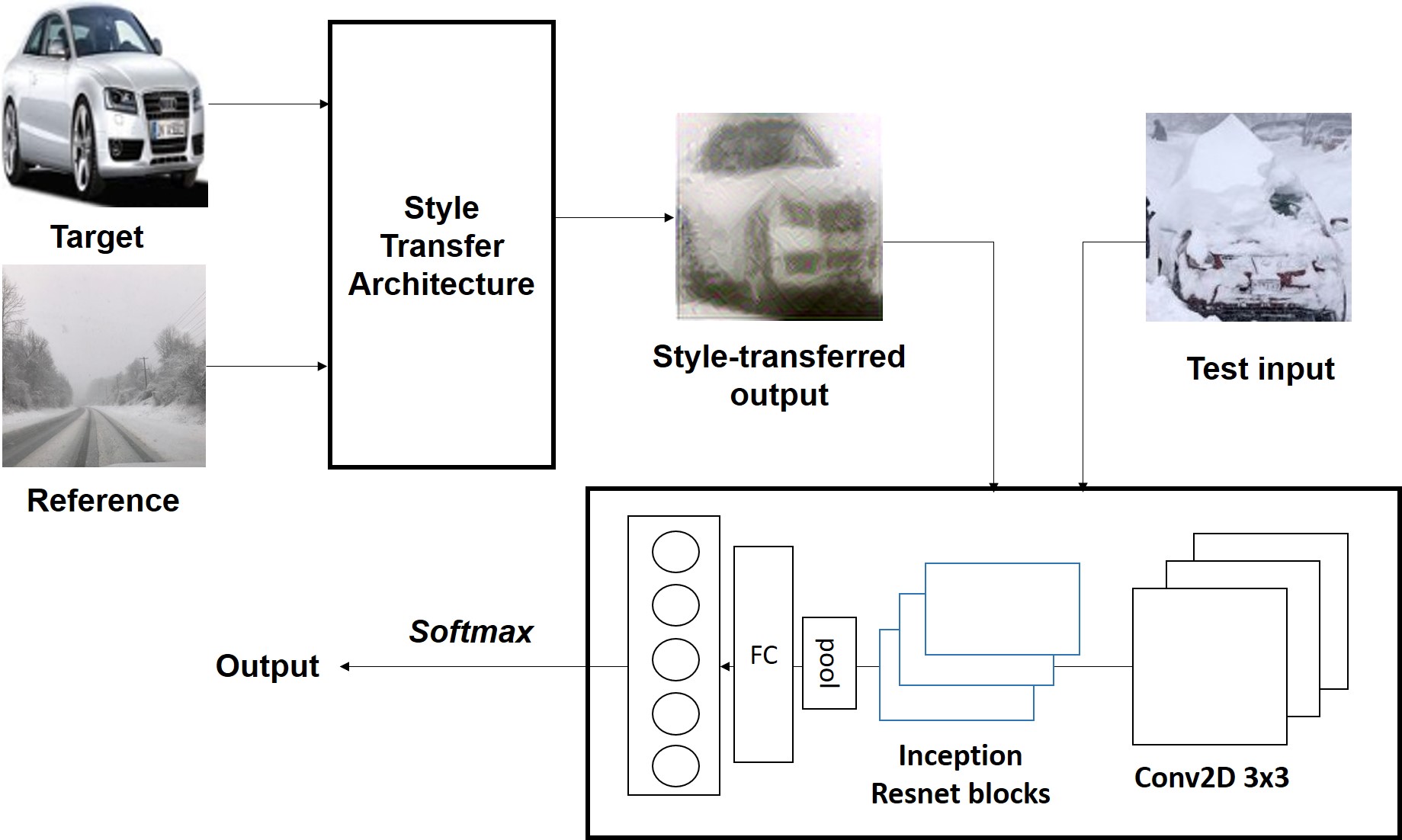}

\end{center}
\caption{ Workflow pipeline of our network architecture. A target and reference image are fed into the style-transfer architecture to generate a composite output. Style-transferred outputs are combined with original car images to form a composite dataset in order to evaluate test set inputs via a InceptionV2-based classifier.
}
\label{fig:fig1}
\end{figure*}
\subsection{Style Transfer}
\label{sec:style}
First defined by Gatys \etal \cite{gatys2015neural} in the context of artwork generation, style transfer refers to the altering of low-level features within an image while retaining its semantic content \cite{gatys2015neural, jackson2018style}. Conceptually, style transfer relies on the underlying fact that different levels of a convolutional neural network capture different aspects of the level. Gatys \etal \cite{gatys2015neural} defined the convolutional features captured by lower levels within the network, such as the local texture or patterns, as the ``style" of an image, while those captured by higher levels as the semantic content of an image. This separate capture process allows for the independent manipulation and transfer of individual components into a generated composite. Gatys \etal's original paper demonstrated how landscape images could be altered to resemble modern artworks while retaining high-level semantic content.

The exact mechanism of style transfer has been discussed extensively in literature \cite{gatys2015neural, li2017demystifying}. Briefly, a VGG network \cite{liu2015very} stripped of any fully connected layers serves as the model for style transfer. To extract high level semantic content, a target image is fed into this network to be encoded by successive convolutional layers. To replicate the features captured at a layer of interest, gradient descent is used on a white-noise image matching feature responses are observed, measured by squared loss error.
\begin{equation}
    L_{content}(\vec{p},\vec{x},l) = \frac{1}{2}\sum_{i,j}(F_{ij}^l - P_{ij}^l)^2
\end{equation}

Where P, F, and l represent the original image, the generated image, and l the layer of interest, respectively. 
To extract the low-level ``style" content of an image, a Gram matrix is constructed to capture the correlations between different filter responses from layer of interest, defined as:

\begin{equation}
    G_{ij}^l = \sum_{k}F_{ik}^l F_{jk}^l
\end{equation}

Or the inner product of the vectorized feature maps i and j within a layer l. Similar to above, gradient descent is then used on a white-noise image to create a matching representation, with the loss defined as the mean-squared distance between the matrices of the original and the generated counterpart.

\begin{equation}
    E_l = \frac{1}{4 N_l^2 M_l^2}\sum_{i,j}(G_{ij}^l - A_{ij}^l)^2
\end{equation}

Finally, an additional loss function is defined that weighs the contribution of each layer to the total loss.
\begin{equation}
    L_{style}(\vec{a},\vec{x}) = \sum_{l=0}^L w_l E_l
\end{equation}

The generated representations can then be combined to form a composite image. Note that the ratio between the two is controlled by manually defined weights, allowing for a degree of control over the outputs. In both the original work and current work, this was done in an iterative manner, so as to evaluate the effect of iterative evolutions of the output images on the classification accuracy.

While style transfer has been extensively demonstrated for generative applications, studies on its usage in a data augmentation context for discriminative applications are limited. Jackson \etal \cite{jackson2018style} demonstrated the capability of neural style transfer in randomizing dataset color, texture and contrast whilst preserving semantic geometry, and improved classification accuracy of vanilla traffic scenes by roughly 1.4\% when compared to traditional data augmentation strategies. Similarly, Zheng \etal \cite{zheng2019stada} observed a 2\% and 1.3\% increases in performance by vanilla VGG16 and VGG19 networks featuring style transfer-based data augmentation on the Caltech 101 and 256 datasets, when compared to models utilizing traditional data augmentation strategies \cite{zheng2019stada}. These works primarily focus on intra-domain discriminative performance improvements, the capability of neural style transfer to improve cross-domain discriminative performance under a real-world context is a topic of interest for research.

%% file: method.tex
\subsection{Data Augmentation using Style Transfer}

Our approach involves the utilization of style transfer to introduce  appropriate domain-specific noise to our vanilla car images in order to create images more reminiscent of adverse winter conditions. Figure \ref{fig:fig1} depicts the overview of our methodology. 

\begin{figure}
\begin{center}
\includegraphics[width=\linewidth]{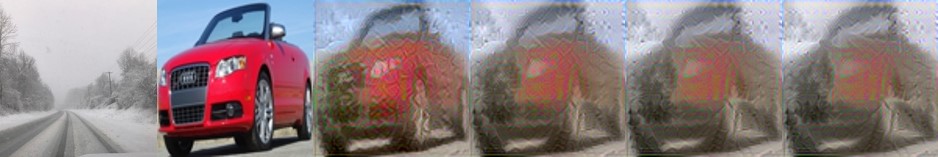}
\end{center}
\caption{ Reference, target, and output images at iterations 1, 3, 5, and 7, used in our experiments.}
\label{fig:fig2}
\end{figure}

Initially, image samples from the original car dataset are fed together with a winter reference image into the VGG19-based neural style transfer architecture to generate synthetic outputs. These outputs are then grouped together with original car images to form a composite dataset representing the generic ``car" class, at a ratio of 20:80.  This dataset was then used to train an InceptionV2-based architecture on the car class and four auxiliary classes. The resulting model is evaluated on a variety of test datasets, and its performance is measured by the true positive (TP) and false positive rates (FP).Figure \ref{fig:fig2} illustrates the style transfer composite generation process in detail. 

As described in Section \ref{sec:style}, a target image is fed into a VGG-19-based convolutional neural network together with a reference winter image. By extracting low-level features from the reference together with high level semantic features from the target image with specific weights, successive style transfer iterations are applied to generate the desired composite output. With an increasing number of iterations, the output becomes correspondingly blurry, as the proportion of low-level features is accumulated. To evaluate the effect of increased iterations, multiple models trained using composite datasets featuring style-transferred images of a specific iteration were systematically evaluated.

%% file: exp.tex
\subsection{Datasets}

We evaluated our method using a composite dataset, based on the natural-images dataset. The natural-images dataset was built by Roy \etal \cite{roy2018effects}, and consists of 8 classes of images from various published datasets \cite{roy2018effects}. For this implementation, only 4 classes of images were considered – car, flower, cat, and dog.

As a test dataset for our class of interest, Google Image Search was used to acquire 200 images of cars under adverse winter conditions, 100 of which were used for training purposes. This dataset (henceforth known as ``Blizzard") consisted of images selected as being human-identifiable as cars: this meant that instances where the vehicle was covered in snow past human legibility were excluded. 
Additionally, 100 images of empty snow landscapes (henceforth known as ``Landscape") were acquired to evaluate the misclassification rates of our models, along with a dataset consisting of 100 style-transferred car images (henceforth known as ``Styled") produced using our approach. Examples of all three datasets are displayed below in Figure \ref{fig:fig3}

\begin{figure}
\begin{center}
\includegraphics[width=\linewidth]{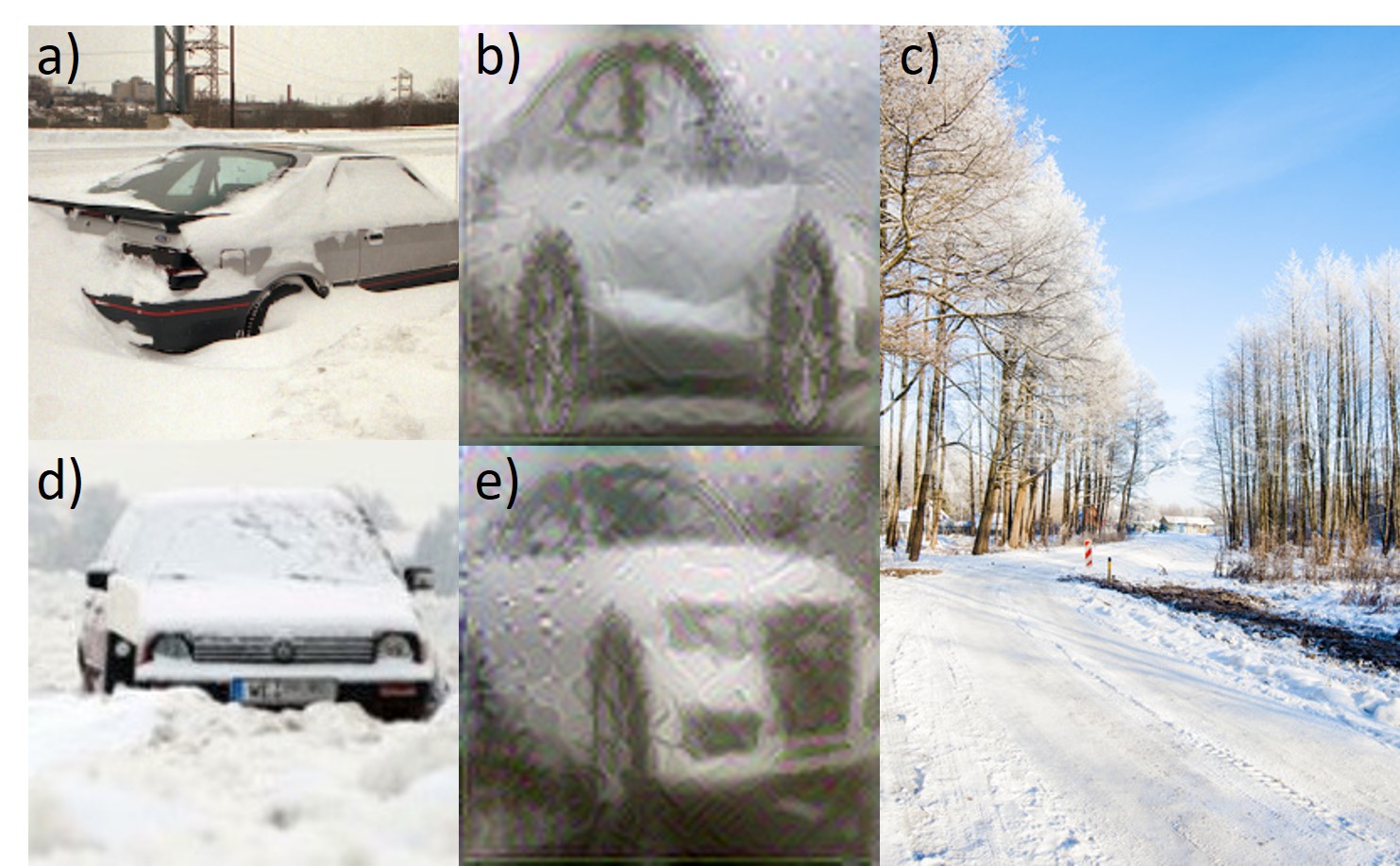}

\end{center}
\caption{Example images of the  ``Blizzard" (a,d),``Styled" (b,e), and ``Landscape" (c)  test datasets used to evaluate our trained models.
}
\label{fig:fig3}
\end{figure}

\subsection{Implementation Details}

The generation of our images was done on a style-transfer convolutional neural network based on the VGG19 architecture as defined by Gatys \etal \cite{gatys2015neural}, with a style weight of 1, content weight of 0.0003, and a total variational weight of 0.00001. For classification,  InceptionV2 ResNet34 architecture \cite{li2017demystifying} was pre-trained for ImageNet classification task. The architecture of the model was modified by removing the output layer prior to the classification layer and include a dropout layer with a dropout rate of 0.5 to limit overfitting during the training process. A small learning rate of 5x$10^{-4}$ was used to fine-tune our model, and the network was retrained on our composite datasets for a maximum of 10 epochs with ADAM optimizer. Traditional augmentation approaches such as image rotation and flipping were also used consistently across all models. Each model was run 20 times to generate statistically significant results. Our code is implemented in Python, using the Keras and Tensorflow packages.

\subsection{Results}

\begin{table*}[!tb]
\begin{center}
\centering
\resizebox{0.8\textwidth}{!}{
\begin{tabular}{|c|c|c|c|}
\hline
TASK & Vanilla(100\%) / A & Styled (20\%) / B & Blizzard (20\%) /C  \\
\hline\hline
Blizzard Images & 0.803$\pm$0.085 & 0.888$\pm$0.035 & 0.967$\pm$0.013 \\
Landscape Images & 0.022$\pm$0.023 & 0.086$\pm$0.060 & 0.236$\pm$0.064\\
Styled Images & 0.102$\pm$0.086 & - & - \\

\hline
\end{tabular}}
\end{center}
\caption{The classification accuracy across our different training domain combinations. Model A was trained with the baseline original car image dataset, while Model B was trained with a augmented dataset, where 20\% of the images in the original dataset were replaced with style-transferred images. Model C was also trained with an augmented dataset, where 20\% of the images were replaced with images from the ``Blizzard" dataset. All three models were then evaluated on the ``Blizzard" and ``Landscape" datasets, respectively, with model A being further tested on the ``Styled" dataset. }
\label{table:table1}
\end{table*}

Table 1 shows the classification accuracy across our different training domain combinations. Model A was trained with the baseline original car image dataset, while Model B was trained with a augmented dataset, where 20\% of the images in the original dataset were replaced with style-transferred images . Model C was also trained with an augmented dataset, where 20\% of the images were replaced with images from the ``Blizzard" dataset. All three models were then evaluated on the ``Blizzard" and ``Landscape" datasets, respectively, with model A being further tested on the ``Styled" dataset. 

The baseline test accuracy utilizing the original car dataset is described by Model A, at \SI{80.3}{\percent} \textpm\ \SI{8.5}{\percent}. The model  possesses the smallest misclassification rate when presented with the landscape data, at \SI{2.2}{\percent} \textpm\ \SI{2.3}{\percent}. The addition of either style transferred or blizzard car images into the training dataset increased the blizzard car detection accuracy while reducing the model uncertainty (from 8.5 to 3.5 and 1.3\%, for vanilla , styled, and blizzard car augmented datasets, respectively). This is supported by the low detection rate (\SI{10.2}{\percent} \textpm\ \SI{8.6}{\percent}) of Model A on the ``Styled" images test dataset, which suggest that the generated data has shifted beyond the original domain. As Model A captures features specific to the original domain, it is difficult for such a model to generalize on out-of-domain data which results in low detection accuracy.  

Interestingly, while Model C certainly possesses the highest detection accuracy and confidence interval among all models, it suffers from high false detection rate when applied to the ``Landscape" dataset, where an increase in rate is 10-fold over Model A (2.2\% to 23.6\%), accompanied by a near 3-fold increase in model-to-model uncertainty (2.3\% to 6.4\%) was observed. The high misclassification rate of Model C may be linked to the high variation found in the real-world dataset images, whether it be difference in background terrain, the weather conditions, or even the amount of snow (noise) observed on the vehicles differing for each image. This is to be expected and remains exceptionally difficult to compensate for due to the levels of entropy observed in weather conditions. It may be possible to improve upon the misclassification rate by the incorporation of a larger dataset, but this may not be feasible.

In contrast, the car images in the ``Styled" dataset are all similar in their level of noise, resulting in a more consistent dataset capable of improving detection accuracy while suffering from only a small increase in misclassification rates, as was observed with Model B. This small increase may be attributed to the randomness observed in the degree of adversity in the conditions of the blizzard car images test set.

Our results are in line with by previous cross-domain work by Jackson \etal \cite{jackson2018style}, who observed a maximum of 11\% increase in classification accuracy for in cross domain applications of the Office dataset when utilizing a combined style-transfer and traditional data augmentation approach compared to purely traditional approach. Notably, the magnitude of the accuracy increase was subject to domain-specific differences between the training data and target data: an accuracy increase of 6.2\% was observed for networks tested on the DSLR domain when trained on the Amazon and Webcam domains.10

Interestingly, when our testing was repeated on models trained on style-transferred outputs generated after 5, 7, or 10, iterations, a progressively worse performance in the composite datasets could be observed, with all models exhibiting notably worse classification accuracy and standard deviation over multiple attempts in comparison to the reference model. We hypothesize the strong stylistic abstraction observed across multiple iterations shifts the domain further from that of the both the vanilla dataset and that of vehicles under adverse conditions, with a detrimental effect on accuracy as a consequence.

%% file: conclusion.tex
In this paper we proposed and evaluated style transfer based data transformation to improve classification accuracy of cars under adverse winter weather conditions. Such augmentation improved the classification accuracy on a foreign domain with only minimal impact on the false detection rate. We compared the performance of our method with the original dataset augmented with real-world blizzard car images. This successfully evaluated our approach to use neural style transfer as being superior to utilizing a purely real-world blizzard dataset, as the latter exhibited high false detection rates due to large variations across its real-world data samples. 

Excessive number of iterations for style transfer may have a detrimental effect on classifier performance. As a rule of thumb, we propose that the appropriate number of iterations of style transfer can be judged visually, and must bear superficial resemblance to the domain of the original dataset.

Given the blur-like qualities of our synthetic generated examples, future studies should focus on the generation of more realistic snow effects to evaluate both the ability of generative models to mimic true reality, and to discern if generated realistic examples of the target domain also suffer from the detrimental effects of high data variation.

Thus, in this paper we evaluate the superiority of a consistent, artificially generated dataset, and show improvements over directly using real-world data given a constrained dataset size. Hence, style transfer is one of the successful ways for synthetic data generation on other foreign domain applications.
